\documentclass[12pt,draftclsnofoot,onecolumn]{IEEEtran}
\usepackage{graphicx}
\usepackage{soul}
\usepackage{booktabs}
\usepackage{multirow}
\usepackage{tikz}
\usepackage{comment}
\usepackage{amsmath,amssymb} 
\usepackage{color}
\usepackage{cleveref}
\usepackage[accsupp]{axessibility}  

\begin{document}
\title{\hspace{-2cm} Learning from Attacks: Attacking Variational\\
\hspace{-2cm} Autoencoder for Improving Image Classification}
\author{Jianzhang Zheng$^1$, Fan Yang$^2$, Hao Shen$^3$, Xuan Tang$^4$, Mingsong Chen$^5$,\\ Liang Song$^6$, Xian Wei$^{7\ast}$\\
{\small zhengjianzhang18@mails.ucas.edu.cn$^1$,}
{\small 208527053@fzu.edu.cn$^2$,}
{\small shen@fortiss.org$^3$,}\\
{\small xtang@cee.ecnu.edu.cn$^4$,}
{\small mschen@sei.ecnu.edu.cn$^5$,}
{\small songl@fudan.edu.cn$^6$,}
{\small xian.wei@tum.de$^{7\ast}$}
\thanks{$^\ast$ Corresponding Author}
}

\maketitle
\begin{abstract}
Adversarial attacks are often considered as threats to the robustness of Deep Neural Networks (DNNs). Various defending techniques have been developed to mitigate the potential negative impact of adversarial attacks against task predictions. This work analyzes adversarial attacks from a different perspective. Namely, adversarial examples contain implicit information that is useful to the predictions i.e., image classification, and treat the adversarial attacks against DNNs for data self-expression as extracted abstract representations that are capable of facilitating specific learning tasks. We propose an algorithmic framework that leverages the advantages of the DNNs for data self-expression and task-specific predictions, to improve image classification. The framework jointly learns a DNN for attacking Variational Autoencoder (VAE) networks and a DNN for classification, coined as Attacking VAE for Improve Classification (AVIC). The experiment results show that AVIC can achieve higher accuracy on standard datasets compared to the training with clean examples and the traditional adversarial training.
\begin{IEEEkeywords}
Adversarial Attacks, Variational Autoencoder, Deep Learning, Image Classification.
\end{IEEEkeywords}
\end{abstract}

\section{Introduction}
Deep Neural Networks (DNNs) have achieved significant success in various applications, e.g. computer vision, natural language processing and speech recognition \cite{otter2020survey_tnnls,voulodimos2018deep_cian,szegedy2013deep_nips,hinton2012deep_spm,covington2016deep_RecSys,krizhevsky2012imagenet_nips}. 
Recent development has shown that DNNs can be vulnerable to well-designed adversarial examples \cite{yuan2019adversarial_nnls,fawzi2018adversarialz_nips,goodfellow2015explaining_iclr}.
Adversarial attacks are considered as the process of generating adversarial examples that mislead DNNs to predict wrong outcome \cite{zhang2021survey_ijcai,chen2020universal_pami,dong2020greedyfool_nips}.
The adversarial attacks are generated by adding crafted perturbations to the benign example that are not perceptible to the human \cite{croce2019sparse_iccv}.
Furthermore, adversarial examples have strong transferability among models \cite{su2018robustness_eccv,wu2018understanding,wiyatno2019adversarial}, which means the adversarial examples generated from one model also have a misleading effect on other models with different architectures.

In order to protect DNNs from adversarial attacks, several strategies including adversarial training \cite{goodfellow2015explaining_iclr,wang2019improving_iclr,zhang2020attacks_icml}, rescaling \cite{guo2017countering_iclr}, denoising \cite{ren2020adversarial,xie2019feature_cvpr} and distillation \cite{papernot2016distillation_sp,goldblum2020adversarially_aaai,liu2019feature_cvpr} were proposed\cite{mao2019metric}.
All these defensive strategies treat the adversarial attacks as threats, and face the problem of degeneration in the prediction performance of benign examples.

Contrary to the view of defending the adversarial attacks, some works \cite{xie2020adversarial_cvpr,ilyas2019adversarial_nips} analyze the adversarial attacks from different perspective. For example, the adversarial attacks carry underlying information hidden in observed data that is beneficial to the task predictions, e.g. image classification.
Due to the transferability of adversarial examples that attack the classifier, classifier-based adversarial examples cannot sufficiently extract features from the data that are helpful for classification.

Recall that most DNNs models are constructed with the aim of either task-specific predictions, e.g. classification and detection, or data self-expression, e.g. reconstruction and generation, the latter can find the explanatory factors that are capable of describing intrinsic structures of the data \cite{bengio2013representation_pami,wei2019trace_pami}.
Therefore, in addition to being task-specific prediction models, DNNs can also be used as self-expressive models. 

Recent research on adversarial attacks on self-expressive models mainly focuses on how to change the input data distribution or the latent representation distribution to generate wrong samples \cite{kos2018adversarial_sp,Tabacof2016Adversarial_arxiv}.
In the field of adversarial learning, self-expressive models and task-specific prediction models can work in cooperation, for example, self-expressive models can also be used to generate adversarial examples to attack classifiers \cite{derui2021man_TDSC,lin2020dual_nips,XiaoL2018Generating_ijcai}, 
or to help the classifier defend against attacks \cite{samangouei2018defense_gan}.
Among the self-expressive models, Variational Autoencoder (VAE) \cite{kingma2014autoencoding_iclr} is a powerful one, which learns the probability distributions of latent variables of input data.
\cite{Tabacof2016Adversarial_arxiv} investigated adversarial attacks for VAE, and found VAE is much more robust to the attack, compared to classifiers.

This paper assumes that the features carried in adversarial examples by attacking the VAE are more useful for classification, compared to classifier-based adversarial examples.
We propose a novel learning paradigm that uses a deep neural network named Generator to learn how to generate adversarial examples that can attack the VAE, and uses these adversarial examples to train a classifier to improve classification accuracy.
The stronger the destructive ability of the adversarial examples generated by the VAE, the better the adversarial examples can capture the key features for improving classification.

\begin{figure}[t]
    \centering
    \includegraphics[width=2.5in]{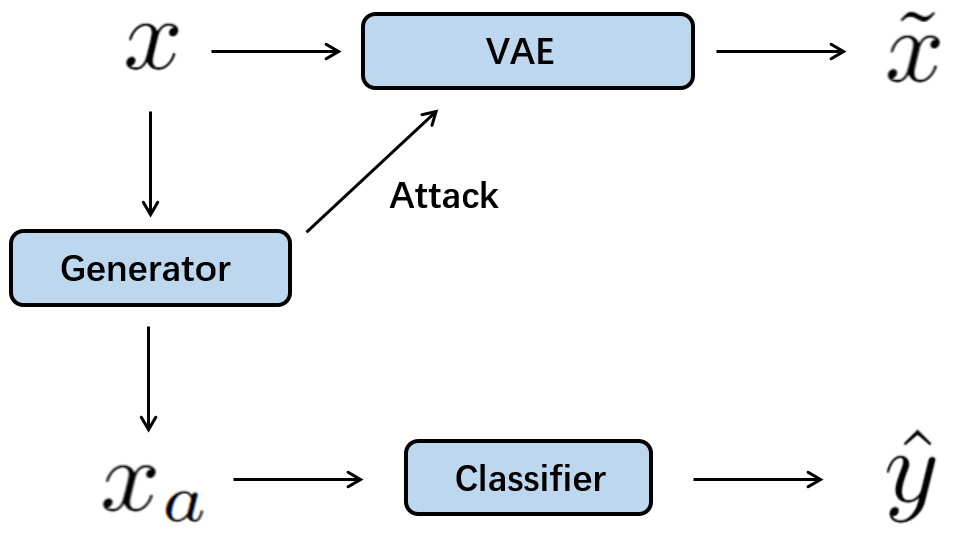}
    \caption{The diagram of AVIC. The notation: $x$ is the original image, ${x}_{a}$ is adversarial examples, $\tilde{x}$ is the reconstruction of $x$, $\hat{y}$ is the prediction of the classifier.}
    \label{framework}
    \vspace{-4mm}
\end{figure} 

In this paper, we propose the AVIC, short for Attacking VAE to Improve Classification.
The proposed joint end-to-end learning diagram of the AVIC consists of four components, as shown in Fig.~\ref{framework}.
%
%
First, we train a VAE on the clean data to let it learn how to extract the structural features of input data.
Secondly, we train a generator to attack the trained VAE via generating the adversarial examples that maximize the loss of VAE. 
After that, we only use the adversarial examples from the generator to train a classifier. 
Finally, by construct a joint end-to-end learning paradigm, the generator and the classifier are globally retrained together. 

The proposed AVIC has following advantages.
\romannumeral 1) The AVIC takes full advantage of 
the unsupervised data self-expression capability of the VAE and 
the supervised learning capability of  classification-driven networks.
Both unsupervised and supervised learning capability are leveraged. 
\romannumeral 2) Since the AVIC can extract the abstract features hidden in the adversarial examples, rather than latent representations of the encoder networks of VAE, the networks for classification are flexible, i.e., we can adopt various classification-driven networks, such as ResNet \cite{he2016deep_cvpr} and VGG \cite{simonyan2015very_iclr}.  
\romannumeral 3) 
The experiment results show that the proposed AVIC can achieve better performance on standard datasets compared to the training
with clean examples and the traditional adversarial training.
%

\section{Related work}
\subsection{Variational Autoencoder}
Variational Autoencoder (VAE) \cite{kingma2014autoencoding_iclr} is a paradigm of self-expressive model, and has a similar structure of autoencoders \cite{hinton1993autoencoders_nips}, which have an encoder and a decoder.
For autoencoders, the encoder learns and compresses the input into a much smaller representation that extracts the most representative information, and the decoder learns how to reconstruct from the compressed representation to the input.
But the encoder of VAE compresses the high-dimensional input into a vector of means and another vector of standard deviations to distribution, and the decoder generates an output similar to the original input based on the latent code sampled from the mean and standard deviation vectors.
Therefore, the VAE has both reconstruction and generation abilities for data.
In addition, it is easier to train VAE, compared to Generative Adversarial Network (GAN) \cite{Goodfell14_GAN_NIPS}, another paradigm of self-expressive models. 


\subsection{Adversarial Examples}
The concept of adversarial examples was first proposed in
\cite{szegedy2013intriguing_iclr}.
\cite{goodfellow2015explaining_iclr} proposed the Fast Gradient Sign Method (FGSM) to generate the adversarial examples and showed that models could be regularized by adversarial training on a mixture of adversarial and clean examples.
\cite{madry2017towards_iclr} proposed Projected Gradient Descent (PGD) which is considered as the strongest first-order attacking, and also proposed another form of adversarial training to only use adversarial examples instead of the mixture of adversarial and clean examples to train models. 
Adversarial examples have different underlying distributions to clean examples \cite{xie2020adversarial_cvpr}, and models trained with adversarial examples would decrease the accuracy on clean images \cite{madry2017towards_iclr}.
In addition to attacking classifiers, adversarial attacks have also been applied to self-expressive models such as VAE and GAN
\cite{kos2018adversarial_sp,Tabacof2016Adversarial_arxiv}.

\cite{ilyas2019adversarial_nips} proposed a new perspective on adversarial examples and categorize the features derived from patterns in the data distribution into robust and non-robust features, where the non-robust features led to the existence of adversarial examples, and proved that non-robust features suffice for standard classification.
\cite{xie2020adversarial_cvpr} proposed added an auxiliary batch norm for adversarial examples in addition to the original batch norm for clean images to reduce the distributions mismatch between clean examples and adversarial examples.

\subsection{Joint Self-expressive and Classification Networks}
Recent works \cite{yoshihashi2019classification_cvpr,robert2018hybridnet_eccv,ghifary2016deep_eccv} have demonstrated the reconstruction networks of self-expressive models can learn the underlying information of input to help image classification, but require the classification networks have the same architectures as the reconstruction networks, which limits their applications.

Since reconstruction networks are useful in representation learning and can learn the underlying information from data without labels, they are often used to self-supervised learning and semi-supervised learning to help the downstream tasks \cite{liu2021self_kde}. 
%
\cite{yoshihashi2019classification_cvpr} performs joint classification and reconstruction on the input data. While  performs classification, where the features from of the intermediate layers the classification network are fed into the reconstruction network to obtain the corresponding latent representation for unknown data detection.
%
\cite{ghifary2016deep_eccv} uses the same parameters in reconstruction network encoder and classification network, and they and they fed the labeled data to the classification part and unlabeled data to the reconstruction part.
\cite{robert2018hybridnet_eccv} adds a reconstruction network branch based on the model proposed by \cite{ghifary2016deep_eccv}. The classification network in the first branch receives supervised signals, and the reconstruction network in the second branch performs unsupervised learning with information that is not useful to the classification network.
Different from these works of training two networks simultaneously, \cite{he2021masked} proposed to first train masked autoencoders for restoring the masked images, and then exploited the encoder of the trained masked autoencoders to extract features directly for classification tasks.

\section{The AVIC Model}
\label{sec:model}
\begin{figure*}[t]
	\centering
	\includegraphics[width = 12cm]{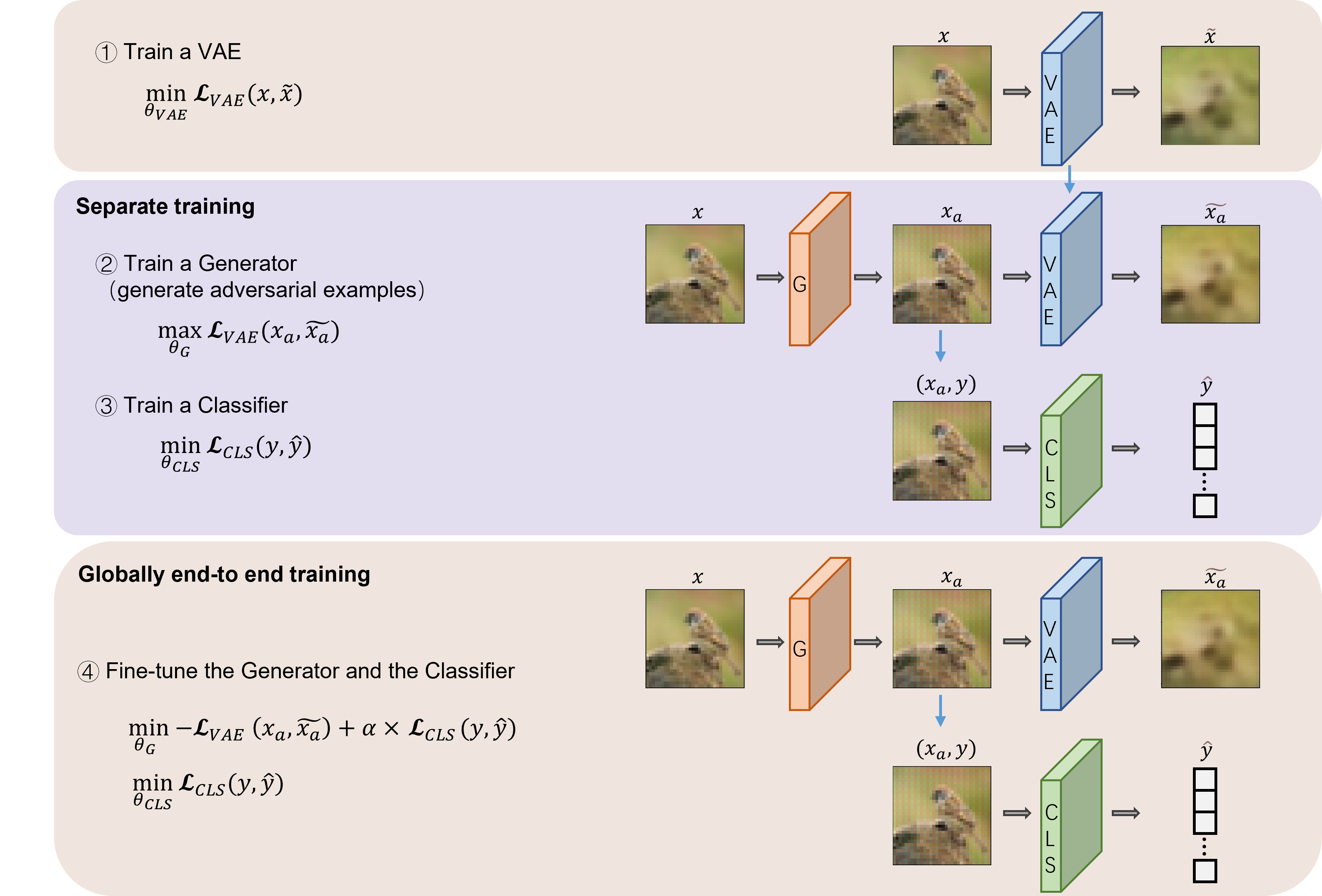}
	\caption{
		The overall pipeline of the proposed \emph{AVIC}. The AVIC contains three stages and four steps. Therein, G denotes the generator and CLS denotes the classifier. Blue arrows indicate that model parameters and image data are passed and left unchanged.
	}
	\label{overview}
	\vspace{-4mm}
\end{figure*}
\subsection{Problem Definition}
The pipeline of the proposed AVIC method is depicted in Fig.~\ref{overview}.
Let $\mathcal{X} \subseteq \mathbb{R}^{H \times W \times C}$  be the input image space, and
$x \in \mathcal{X}$ be a sample image. 
A VAE network is firstly trained to learn the map $f_{VAE}:\mathcal{X} \to \mathcal{X}$, 
so that $\widetilde{x} := f_{VAE}(x)$ is an approximate of $x$.
Then, a generator is trained to produce the adversarial examples $x_{a}$ based on $x$, 
which can make the reconstruction error, i.e., the difference between $\widetilde{x}_{a}$ and $x_{a}$ as large as possible.
After that, $x_{a}$ from the trained generator is provided to train the classifier.
Finally, fine-tune the generator and the classifier jointly.

\subsection{Training of VAE}
Let us assume that the input image $x \in \mathcal{X} $ is sampled from the intractable real distribution $p_r(x)$. The decoder of VAE learns a distribution $p_{\theta}(x)$ for approximating $p_r(x)$.
Let $\phi$ be the parameters of the encoder, and $q_{\phi}(z | x)$ denotes the conditional probability of the encoder producing the latent representation $z$ based on the input $x$. 
Similarly, let $p_{\theta}(x | z)$ be the distribution of $x$ generated by the decoder given the latent code $z$ from the encoder.
%

For the purpose of minimizing the discrepancy between the produced images and the original images, we train the VAE model $f_{VAE}$ by maximizing the Evidence Lower BOund (ELBO) \cite{kingma2014autoencoding_iclr}, which can be expressed as:
\begin{equation}
\begin{aligned}
{\arg \max_{\theta, \phi}}~{\rm ELBO} \left(\theta, \phi ; x\right)= \mathbb{E}_{q_{\phi}\left(z | x\right)}\left[\log p_{\theta}\left(x | z\right)\right]
-D_{K L}\left(q_{\phi}\left(z | x\right) \| p_{\theta}(z)\right).
\end{aligned}
\label{VAE}
\end{equation}
The loss function of VAE is given as in Eq.~\eqref{l_vae}, which consists of the reconstruction error and the KL-divergence, i.e.,
\begin{equation}
\begin{aligned}
\min \mathcal{L}_{V A E}(x, \widetilde{x})&={\arg \min_{\theta, \phi}} \ {\rm -ELBO} \left(\theta, \phi ; x\right)\\
&=\frac{1}{2} (x-\widetilde{x})^2 + D_{K L}\left(q_{\phi}\left(z | x\right) \| p_{\theta}(z)\right).
\label{l_vae}
\end{aligned}
\end{equation}
The first item drives the output restored images $\widetilde{x}$ as close as possible to the original images $x$, and the later is a regularization item, which makes the posterior distribution $q_{\phi}\left(z | x\right)$ approximate to the prior distribution $p_{\theta}(z)$. 

In next subsection, we will introduce a joint training paradigm to train attacking and classification networks.

\subsection{Joint Training of Attacking and Classification}
The joint training paradigm consists of two stages. 
The first stage is a separate training step, which includes attacking the trained VAE to train a generator to generate adversarial examples, and then using the adversarial examples from the trained generator to train the classifier.
The second stage is a globally end-to-end training step, which combines the processes of attacking the VAE and training classifier to achieve global optimization. 

\begin{figure}[t]
    \centering
    \includegraphics[width=12cm]{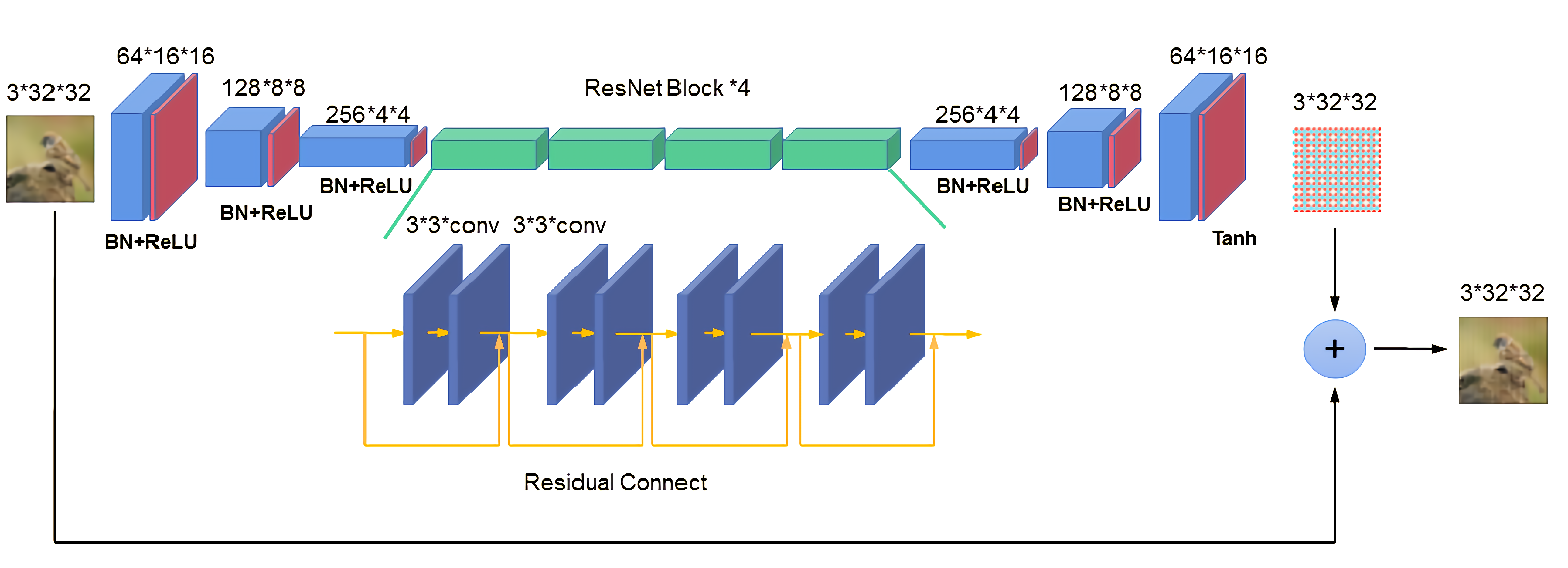}
    \caption{Illustration of the architecture of the generator.}
    \label{Generator}
    \vspace{-4mm}
\end{figure}

\textbf{Separate Training.} 
Adversarial examples are considered carrying the underlying features helpful for the supervised tasks.
To generate attacks that degrade the reconstruction quality of the VAE, we introduce a DNNs-based generator $f_{\theta _G}$ to produce the adversarial example $x_a$ given the clean input $x$.
The architecture of generator is shown in Fig.~\ref{Generator}, which consists of three parts: encoder, bottleneck, and decoder.

Wiht the parameters of the VAE fixed, we train the generator to attacking the VAE via maximizing the $\mathcal{L}_{V A E}$.
The objective function of the generator during separate training is written as,
\begin{equation}
\begin{aligned}
\mathcal{J}_{separate}(G) = {\arg \ \max_{\theta _G}} \mathcal{L}_{V A E}(x_a, \widetilde{x}_{a}) = {\arg \ \max_{\theta _G}} \mathcal{L}_{V A E}(f_{\theta _G}(x), \widetilde{f_{\theta _G}}(x)),
\end{aligned}
\label{attack}
\end{equation}
where $\widetilde{x}_{a}$ is the reconstruction of $x_a$ by the VAE.

After the generator finishes training, we turn to train the classifier.
We feed the clean images $x$ to the generator to generate adversarial examples $x_a$, and use $x_a$ and the label $y$ corresponding to $x$ to train classifier,
\begin{equation}
\begin{aligned}
{\arg \min_{\theta _{CLS}}}  \mathcal{L}_{CLS}
&={\arg \min_{\theta_{CLS}}}{\rm Cross Entropy Loss}(y,\hat{y})\\
&= {\arg \min_{\theta_{CLS}}}{\rm Cross Entropy Loss}(y,f_{\theta _{CLS}}(x_a)).
\label{loss_cls}
\end{aligned}
\end{equation}


\textbf{Globally end-to-end Training.} 
To achieve higher recognition accuracy, we add the classification loss function to the objective function of the generator. 

\begin{equation}
\begin{aligned}
&\mathcal{J}_{global}(G) ={\arg\min_{\theta_{G}}} -\mathcal{L}_{V A E}\left(x_{a}, \widetilde{x}_{a}\right) +\alpha \times \mathcal{L}_{C L S}(y, \hat{y}) \\ 
&={\arg\min_{\theta_{G}}} -\mathcal{L}_{V A E}\left(f_{\theta _{G}}(x), \widetilde{f_{\theta _{G}}}(x)\right) +\alpha \times \mathcal{L}_{C L S}(y, f_{\theta _{CLS}}(f_{\theta _{G}}(x))),
\label{global}
\end{aligned}
\end{equation}
where $\alpha$ weighs $\mathcal{L}_{C L S}$ against $\mathcal{L}_{V A E}$.
The parameters of the classifier will be updated, but the objective function of the classifier during global training is sill in the form of Eq.~\eqref{loss_cls}.

Different from adversarial attack methods such as FGSM and PGD, the proposed generator-based attacking strategy is able to combine the processes of adversarial attack and classification training for eliminating the redundant or non-discriminative features, which reduce the performance of the supervised learning in the adversarial examples. 
For another reason, the global training makes the learned inner- and inter-signal representations more compatible and targeted for the supervised tasks. 



\section{Experiments}
In this section, we first introduce the implementation details in \cref{subsec:setup}, including the image datasets, the networks architecture of models and the parameter setups for training models. 
Then, in order to verify the effectiveness of
the proposed method 
on image classification tasks, we compare the performance of classifiers trained with clean data without attacks, adversarial attacks on classification and adversarial attacks on VAE, respectively.
After that, ablation studies about the architectures of classifier, attacker strength and the cost function of global training are performed in \cref{subsec:e2}.
Finally, we further study the influence of adversarial attacks for VAE on robustness of classifier, and compare with that of adversarial attacks for classifier in \cref{subsec:e3}.

\subsection{Implementation Details}
\label{subsec:setup}
\textbf{Datasets.} Our method is validated by experiments on four benchmark image datasets including MNIST \cite{lecun1998_mnist}, CIFAR10\cite{krizhevsky2009learning_cifar10_100} and CIFAR100\cite{krizhevsky2009learning_cifar10_100}. 
%
The evaluation metric of classifiers is the standard testing accuracy on original clean test sets of these benchmark image datasets.

\begin{figure}[t]
    \centering
    \includegraphics[width=12cm]{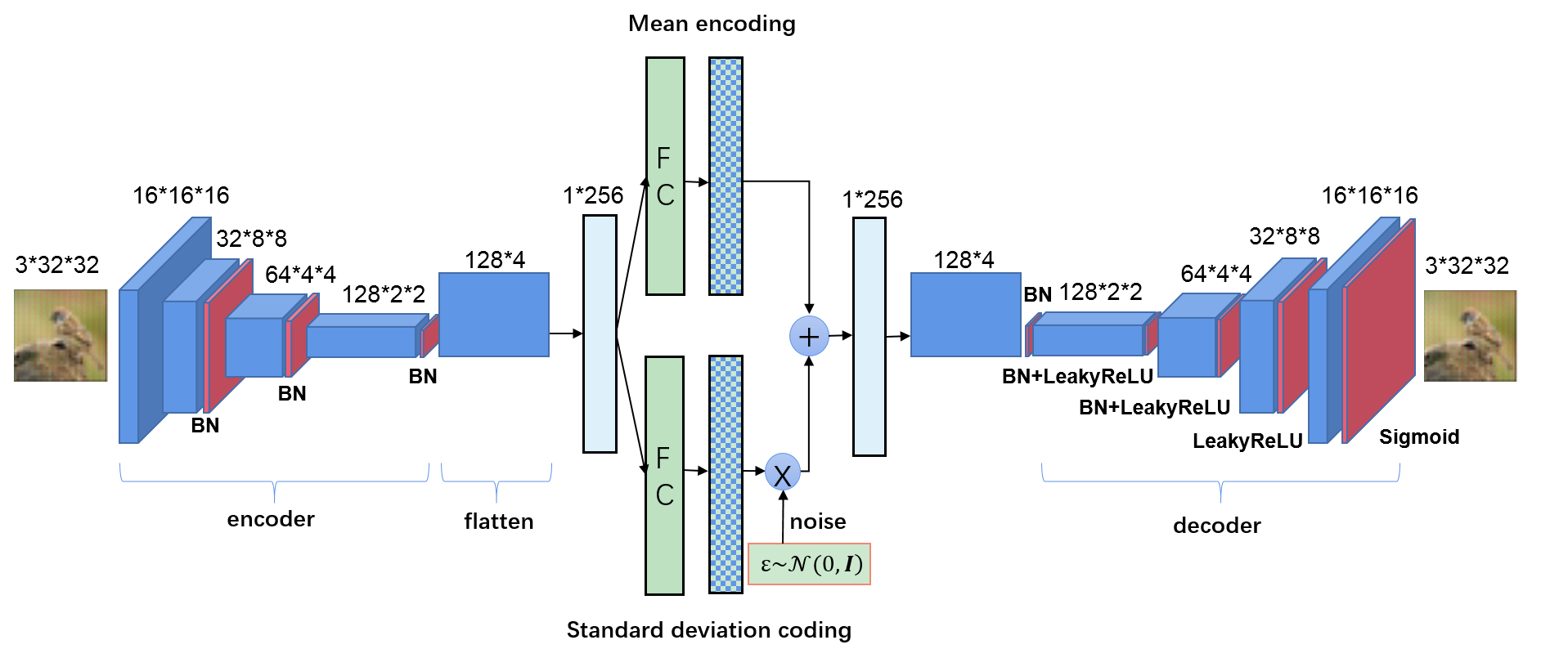}
    \caption{Illustration of the architecture of the VAE.}
    \label{arch_VAE}
\end{figure}

\textbf{Model architectures.} 
As discussed previously, the proposed AVIC consists of three parts, i.e., VAE, generator and classifier.
The generator we used has been shown in Fig.~\ref{Generator}.
The architecture of the VAE in AVIC is shown in Fig.~\ref{arch_VAE}, which is a modified version on the basis of the original VAE \cite{kingma2014autoencoding_iclr}.

The classifier that we adopted in most cases is the ResNet-18 \cite{he2016deep_cvpr}, which is used to measure the quality of different data within classification tasks. In this work, the classifier trained with CIFAR10 is regarded as the standard classification model if it is not specially specified. 
In addition, VGG-19 \cite{simonyan2015very_iclr} is introduced as a supplementary model in ablation experiments to explore the effects of model architectures.
VGG-19 is also used as a reference network when analyzing the influence of different classification networks. 
To produce the adversarial examples based on the original images, a generator whose output can disturb the reconstruction of the VAE is designed.

\textbf{Parameter setups for training.} The value range of all the image samples is rescaled into $[0,1]$. To limit the noise from the generator within an upper bound, we impose the infinite norm $l_{\infty}$ constraint to the noise. 
%
In all experiments without indicating the perturbation size $\epsilon$, the value of $\epsilon$ is taken as 0.02.
For training the ResNet and the VAE, we set the learning rate to 0.005 and 0.05 respectively and adopt the Adam optimizer.

\subsection{Adversarially Attacking VAE for Improving Classification}
\label{subsec:e1}
\begin{figure}[tb]
    \centering
    \includegraphics[width=3 in]{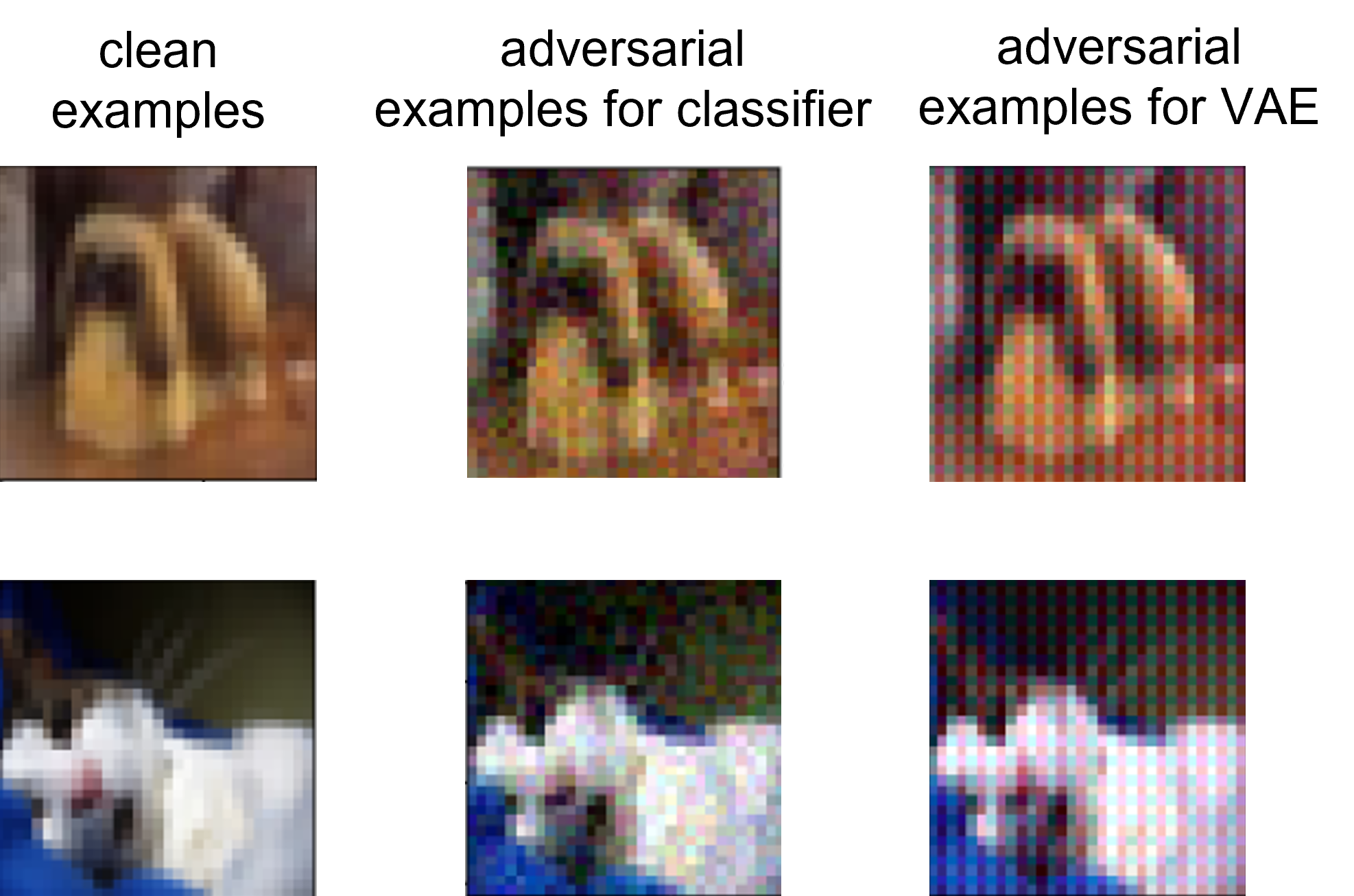}
    \caption{Illustration of clean examples, adversarial examples for classifier and adversarial examples for VAE (from left to right). These images are sampled from CIFAR10 and the perturbation size of adversarial examples is set to be 0.1}
\label{AR_AC}
\end{figure}

\begin{figure}[tbp]
    \centering
    \includegraphics[width=3in]{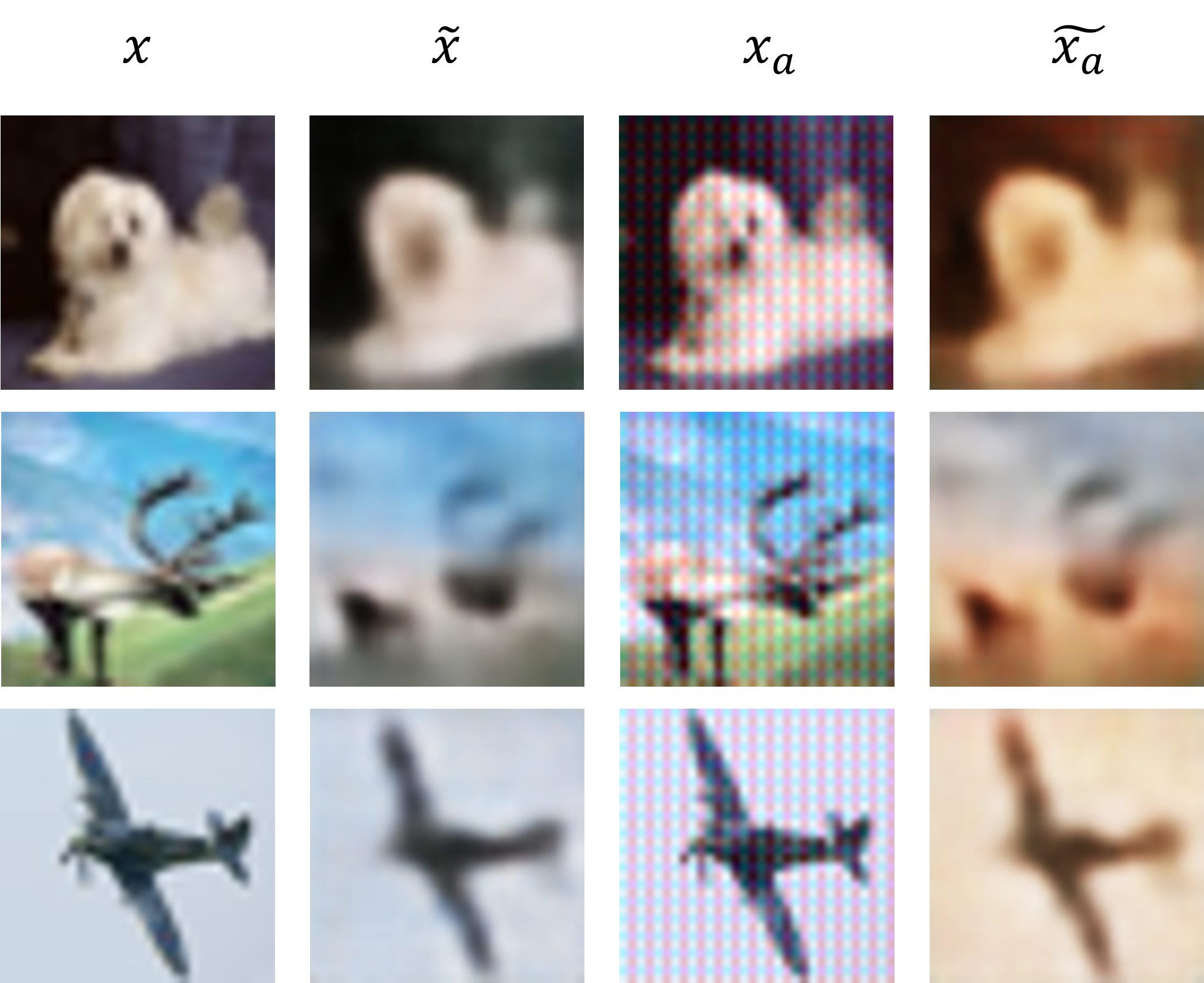}
    \caption{Illustration of original images, the reconstruction of original images, adversarial examples on VAE and the reconstruction of adversarial examples (from left to right).}
\label{x_xr_xa_xar}
\end{figure}

In this subsection, the performance of our proposed adversarial examples on VAE are compared with clean examples without attacks and adversarial examples on classifier.
We compare the performance of classifiers trained with three types of data: (1) clean examples unperturbed by adversarial attacks, (2) adversarial examples produced by adversarial attacks on the classification network, and (3) adversarial examples produced by adversarial attacks on the VAE. 
The third one is in the framework of AVIC. 
Some examples of the above three types of images are shown in Fig.~\ref{AR_AC} .
\begin{table}[tbp]
  \centering
  \caption{Accuracy of models trained with clean examples $x$ and adversarial examples $x_a$ respectively on various datasets. In the way of training the model with $x_a$, ${\rm AVIC_{w/o \; global}}$ denotes the AVIC without global training.  
  }
    \begin{tabular}{ccccc}
    \toprule
     Input     & Setting       & MNIST & CIFAR10 & CIFAR100 \\
    \midrule
    $x$ & vanilla training & 99.64 & 83.00    & 62.76 \\
    \midrule
    \multirow{2}[2]{*}{$x_a$} & ${\rm AVIC_{w/o \; global}}$ & 99.65 & 83.15 & 63.65 \\
          & AVIC & 99.70  & 84.34 & 63.65 \\
    \bottomrule
    \end{tabular}%
  \label{datasets}%
    \vspace{-4mm}
\end{table}%
\begin{table}[tbp]
  \centering
  \caption{Accuracy of classifiers trained with different examples generated in different ways. The column of 'attack' d different ways of generating adversarial examples. 'G-s' and 'G-g' denote methods based on generator with separate training and global training respectively. 'AC' and 'AR' mean adversarial attacks on classification and reconstruction respectively.}
    \begin{tabular}{cccccc}
    \toprule
    \multirow{2}[4]{*}{} & \multicolumn{4}{c}{attack}   & no attack \\
\cmidrule{2-6}          & FGSM  & PGD   & G-s   & G-g   & clean examples \\
    \midrule
    AC    & 71.61 & 77.91 & 79.04 & 80.38 & \multirow{2}[2]{*}{81.70} \\
    AR    & 82.68 & 82.94 & 82.34 & \textbf{84.31} &  \\
    \bottomrule
    \end{tabular}%
  \label{tab:attack}%
    \vspace{-4mm}
\end{table}%
Among them, The reconstruction of adversarial examples and the reconstruction of clean examples for VAE are compared in Fig.~\ref{x_xr_xa_xar}. 
From the comparison in Fig.~\ref{x_xr_xa_xar}, we find that after adding adversarial perturbations to the clean image, the reconstructed image obtained by the same VAE is quite different from the reconstruction of the original image, which is mainly reflected in the color of the images.
We denote the training setting that only uses clean examples as vanilla training.
As shown in Table~\ref{datasets} and Table~\ref{tab:attack}, the models trained with adversarial examples for VAE achieve higher classification accuracy on the testing set than the model trained with clean examples.
What's more, the changes of adopted datasets and ways of generating adversarial examples have no impact on the accuracy.
These results demonstrate the adversarial examples that are generated by adversarial attacks on VAE contain more useful features than original examples. 
In other words, the classifiers indeed learn more discriminative information from the adversarial attacks.

On the other hand, adversarial examples for classifier and VAE respectively are also compared in Fig.~\ref{AR_AC}. 
It shows that perturbations of different forms from adversarial examples for classification are added into adversarial examples for reconstruction. %
The former perturbations are disorganized, while the latter are regular and grid-like. Combining the results in Table~\ref{tab:attack}, it's obvious that attacks for reconstruction networks can improve classification, while attacks for classification degrade the performance of image discrimination on the contrary. 

\begin{table}[tbp] 
  \centering
  \caption{Accuracy of models trained with adversarial examples with different perturbation size on difficult datasets.}
    \begin{tabular}{ccccc}
    \toprule
    \multicolumn{2}{c}{perturbation size $\epsilon$} & MNIST & CIFAR10 & CIFAR100 \\
    \midrule
    0 & clean & 99.64 & 83.00    & 62.76 \\
    \midrule
    \multirow{2}[2]{*}{0.01} & separate & 99.65 & 83.15 & 63.64 \\
          & global & 99.70  & 84.34 & 63.65 \\
    \midrule
    \multirow{2}[2]{*}{0.03} & separate & 99.62 & 82.22 & 62.97 \\
          & global & 99.66 & 83.59 & 63.07 \\
    \midrule
    \multirow{2}[2]{*}{0.05} & separate & 99.64 & 80.79 & 62.79 \\
          & global & 99.66 & 82.15 & 62.95 \\
    \midrule
    \multirow{2}[2]{*}{0.07} & separate & 99.67 & 79.20  & 59.96 \\
          & global & 99.62 & 80.16 & 60.00 \\
    \midrule
    \multirow{2}[2]{*}{0.09} & separate & 99.65 & 78.14 & 58.86 \\
          & global & 99.68 & 79.06 & 58.89 \\
    \bottomrule
    \end{tabular}%
  \label{ep3}%
    \vspace{-4mm}
\end{table}%

\subsection{Ablation Studies}
\label{subsec:e2}
This subsection is to verify the necessity of each module in the proposed AVIC. 

\textbf{Adversarial attacker strength.} The adversarial attacker strength corresponds to the sizes of adversarial perturbations $\epsilon$ added to clean images. 
The results in Table~\ref{ep1} demonstrate that the models trained with the adversarial examples with different perturbation intensity mostly achieve lower standard accuracy than the standard classification models.
\begin{table}[tbp] 
  \centering
  \caption{Accuracy of ResNet trained with adversarial examples with different perturbation size.}
    \begin{tabular}{cc}
    \hline
    perturbation size $\epsilon$ & ResNet \\
    \hline
    0     & 81.70 \\
    0.03  & 83.22 \\
    0.1   & 64.79 \\
    0.3   & 66.63 \\
    0.8   & 28.47 \\
    \hline
    \end{tabular}%
  \label{ep1}%
\end{table}%
However, when $\epsilon=0.03$, 
the result becomes reverse. And the same conclusion is also drawn when the perturbation size is between 0.1 and 0.5, as shown in Table~\ref{ep2}. 

\begin{table}[tbp] 
  \centering
  \caption{Accuracy of ResNet and VGG trained with adversarial examples with different perturbation size.}
    \begin{tabular}{ccc}
    \hline
    perturbation size $\epsilon$ & ResNet & VGG \\
    \hline
    0     & 81.70  & 86.28 \\
    0.01  & 83.44 & 86.45 \\
    0.02  & 84.31 & 86.60 \\
    0.03  & 83.22 & 86.66 \\
    0.04  & 81.95 & 86.69 \\
    0.05  & 81.55 & 86.77 \\
    \hline
    \end{tabular}%
  \label{ep2}%
\end{table}%

In addition, experiments with different perturbation sizes are also conducted on MNIST, CIFAR10 and CIFAR100 datasets and the results are shown in Table~\ref{ep3}.
In summary, in the condition of large adversarial perturbation, the adversarial examples damage the classification accuracy. This is because strong adversarial attacks destroy the intrinsic information that can help classification in clean examples. However, when the value of $\epsilon$ is small, the adversarial examples generated by our method can improve the classifier performance better than the clean examples no matter what classification network is tested. 
%
%
%
\begin{table}[tbp] 
  \centering
  \caption{Accuracy of classifiers trained with difference $\alpha$ in global training on different datasets. $\alpha = 0$ means the generator is trained only with $\mathcal{L}_{V A E}$ and the classifier is still trained with $\mathcal{L}_{C L S}$.}
    \begin{tabular}{cccc}
    \toprule
    $\alpha$ & MNIST & CIFAR10 & CIFAR100 \\
    \midrule
    0     & 99.65 & 83.91 & 63.65 \\
    0.001 & 99.65 & 83.96 & 63.65 \\
    0.01  & 99.65 & 83.93  & 63.65 \\
    0.1   & 99.65 & 83.89 & 63.65 \\
    1     & 99.65 & 83.98 & 63.65 \\
    10    & 99.65 & 84.06  & 63.65 \\
    100   & 99.65 & 83.89 & 63.65 \\
    1000  & 99.65 & 83.95 & 63.65 \\
    \bottomrule
    \end{tabular}%
  \label{alpha}%
\end{table}%
\begin{table}[tbp] 
  \centering
\caption{Accuracy of ResNet and VGG trained with examples generated by different attack methods. 'no attack' correspond clean examples. 'G-s' and 'G-g' denote methods based on generator with separate training and global training respectively.}
    \begin{tabular}{ccc}
    \hline
    attack methods & ResNet & VGG \\
    \hline
    no attack & 81.70  & 86.28 \\
    FGSM  & 82.68 & 86.35 \\
    PGD   & 82.94 & 86.60 \\
    G-s   & 82.34 & 86.60 \\
    G-g   & 84.31 & 87.91 \\
    \hline
    \end{tabular}%
  \label{model}%
    \vspace{-4mm}
\end{table}%

\textbf{Methods of generating adversarial examples.} Generally, it's by iterative or optimized methods that adversarial examples are generated from original images. In contrast, the way that we propose to generate adversarial examples is utilizing a generator and thus can perform global optimization by simultaneously training a generator and a classifier. 
Results shown in Table~\ref{tab:attack} reveal that the adversarial examples generated by a separate trained generator or common adversarial attack methods like FGSM and PGD are equally effective in helping classification tasks. However, results have changed after adding global training for fine-tuning the generator, the classification accuracy of the obtained model is about 2\% higher than that of the above-mentioned methods. It's likely that global training enables the redundant or non-discriminative parts of the generated adversarial examples to be further eliminated. After removing those distracting information in the images, the classification model is better able to learn useful features and thus achieve higher accuracy. This presents the evidence for that our proposed method can better learn from the attack for reconstruction to improve classification.

\textbf{Cost function in global training.} In Eq.~\eqref{global}, there is a hyperparameter $\alpha$ in the cost function of global training. The parameter $\alpha$ only influences the generator. When $\alpha=0$, the effects of adversarial attack are only considered when training the generator. As $\alpha$ increases, the weight of helping classification in cost function gradually increases. As a result, $\alpha$ represents the trade-off between boosting adversarial attacks and improving classification when training the generator.
Table~\ref{alpha} shows that different values of $\alpha$ lead to similar performances of corresponding classifiers on three different datasets. On MNIST database, the accuracy in different $\alpha$ is even the same.
%



\textbf{Architectures of classifier.} Different architectures of classifier are adopted to explore the influence on our proposed method. The results in Table~\ref{ep2} and Table~\ref{model} suggest that the network structure doesn't affect the effect of our method, that is, it can also improve the accuracy of the classifier whatever structure the classifier consists of.

\subsection{The Effects of Adversarial Examples on Different Models}
\label{subsec:e3}
\begin{figure}[tbp]
    \centering
    \includegraphics[width=3in]{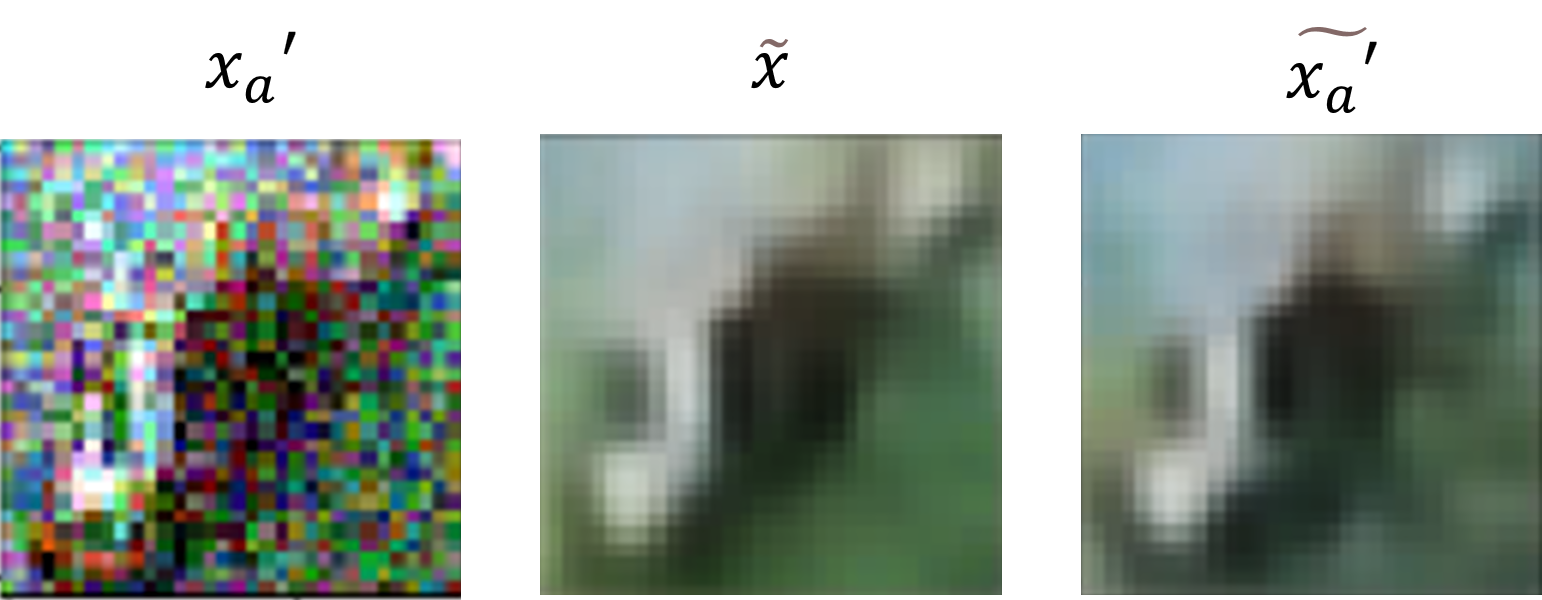}
    \caption{Illustration of an adversarial example for classifier, the reconstruction of clean image and the reconstruction of the adversarial example. }
\label{AC_recon}
  \vspace{-4mm}
\end{figure}

\begin{table}[tbp]
  \centering
  \caption{The effects of adversarial examples for VAE and classification on both two models. In the table, PGD represents adversarial examples generated by PGD for classification. The larger the increasing rate is, the more successful the attack is.}
    \begin{tabular}{ccccc}
    \toprule
    \multicolumn{2}{c}{\multirow{2}[4]{*}{Input}} & \multirow{2}[4]{*}{Classification Accuracy(\%)} & \multicolumn{2}{c}{VAE Loss} \\
\cmidrule{4-5}    \multicolumn{2}{c}{} &       & Value & Increasing Rate(\%) \\
    \midrule
    \multicolumn{2}{c}{Clean Examples} & 99.80/81.15 & 1.73 & - \\
    \midrule
    \multirow{4}[2]{*}{PGD} & e=0.03 & 0.17/10.27 & 1.73  & 0.12 \\
          & e=0.1 & 0.08/6.23 & 1.80 & 0.56 \\
          & e=0.3 & 0.04/2.48 & 1.78 & 6.13 \\
          & e=0.8 & 0.03/1.62 & 2.02 & 16.41 \\
    \midrule
    \multirow{4}[2]{*}{VAE} 
          & e=0.03 & 83.00/70.00 & 1.72 & 1.17 \\
          & e=0.1 & 13.36 & 1.885 & 8.33 \\
          & e=0.3 & 10.80/10.25 & 2.33 & 33.09 \\
          & e=0.8 & 10.00    & 2.78 & 60.71 \\
    \bottomrule
    \end{tabular}%
  \label{tab:rc}%
\end{table}%
Experiments in Table~\ref{tab:rc} display the effects of adversarial examples for reconstruction and classification on both two models. 
PGD adversarial examples can significantly reduce the accuracy of the classification model from about 80\% to around 0.17\%, while the influence of their attacking the VAE is limited with the highest increasing rate of VAE loss 16.41\%. 
These results show that the adversarial examples can easily destroy the function of the classification networks, but the damage to the reconstruction network is relatively small. The Fig.~\ref{AC_recon} shows that the reconstructed image of adversarial example for classification and clean example are similar in the aspects of shape, outline and color. For the human being, there is almost no difference between two images. 
While different from that, the reconstructed image of the adversarial example for reconstruction has a large change in color from the reconstructed clean image. 
As for adversarial examples for reconstruction, they can also degrade the performance of classifiers. When the perturbation intensity is relatively large, the accuracy drops a lot. However, the reconstruction adversarial examples just degrade the accuracy by 10\% in the case that others degrade by about 80\%. This might be the foundation of using adversarial examples to improve classification accuracy.

\section{Conclusions}
In this paper, we proposed an algorithmic framework called AVIC, which treated adversarial attacks from a positive perspective and leveraged them to improve accuracy of image classification task.
The AVIC includes a VAE, a generator and a classifier.
The VAE learns how to characterize the underlying distribution information of image space. 
The generator learns to attack the trained VAE in order to let the generated adversarial examples can extract abstract representations that improve the training of the classifier.
We evaluate the AVIC and the models trained with original images across various benchmark datasets.
The comparison results reveal that the classifier under the framework of AVIC achieves higher accuracy.
In addition, we conduct ablation studies to verify the effectiveness of the AVIC, and conduct experiments to show that adversarial attacks can not transfer between the classifier and VAE networks.
The architecture of the proposed AVIC is flexible and can be extended to more general cases of attacking self-expressive DNN models, such as general autoencoders and self-supervised learning models, for improving classification. 
%
%

\bibliographystyle{IEEEtran}
\bibliography{learn_from_attacks}
\end{document}